\pdfoutput=1

\documentclass[11pt]{article}

\usepackage[final]{acl}

\usepackage{times}
\usepackage{latexsym}
\usepackage{multirow}
\usepackage[export]{adjustbox}
\usepackage{amsfonts}
\usepackage{graphicx}
\usepackage{booktabs}
\usepackage{xcolor}
\usepackage{caption}
\usepackage{subcaption}

\usepackage[T5]{fontenc}

\usepackage[utf8]{inputenc}

\usepackage{microtype}

\usepackage{inconsolata}

\usepackage{graphicx}

\usepackage{amsmath}

\usepackage{pifont}
\newcommand{\cmark}{\ding{51}}%
\newcommand{\xmark}{\ding{55}}%

\title{ViConsFormer: Constituting Meaningful Phrases of Scene Texts using Transformer-based Method in Vietnamese Text-based Visual Question Answering}

\author{
 \textbf{Nghia Hieu Nguyen\textsuperscript{1,3}},
 \textbf{Tho Thanh Quan\textsuperscript{1,3}},
 \textbf{Ngan Luu-Thuy Nguyen\textsuperscript{2,3}}
\\
 \textsuperscript{1}Ho Chi Minh city University of Technology, \\
 \textsuperscript{2}University of Information Technology, \\
 \textsuperscript{3}Vietnam National University, Ho Chi minh city, Vietnam, \\
 \texttt{\{nhnghia.sdh231,qttho\}@hcmut.edu.vn, ngannlt@uit.edu.vn}
\\
 \small{
   \textbf{Correspondence:} \href{mailto:qttho@hcmut.edu.vn}{Tho Thanh Quan}, \href{mailto:ngannlt@hcmut.edu.vn}{Ngan Luu-Thuy Nguyen}
 }
}

\begin{document}
\maketitle
\begin{abstract}
Text-based VQA is a challenging task that requires machines to use scene texts in given images to yield the most appropriate answer for the given question. The main challenge of text-based VQA is exploiting the meaning and information from scene texts. Recent studies tackled this challenge by considering the spatial information of scene texts in images via embedding 2D coordinates of their bounding boxes. In this study, we follow the definition of meaning from linguistics to introduce a novel method that effectively exploits the information from scene texts written in Vietnamese. Experimental results show that our proposed method obtains state-of-the-art results on two large-scale Vietnamese Text-based VQA datasets. The implementation can be found at this \href{https://github.com/hieunghia-pat/ViConsFormer}{link}.
\end{abstract}

\section{Introduction}

\label{sec:intro}

Multimodal learning, particularly vision-language tasks, has recently attracted the attention of the research community. Visual Question Answering (VQA) \cite{VQA} is one of the well-known tasks in vision-language studies. This task gives the machines a question and an image. The machines are required to find the evidence in the image to answer the given question.

Text-based VQA \cite{singh2019towards} is an advanced version of the VQA task in which, besides the visual information in the images, the machines are required to incorporate the information of scene texts for more accurate answers.

Various datasets were constructed for researching Text-based VQA tasks, especially in high-resource languages such as English \cite{VQA,Goyal2016MakingTV,singh2019towards,Biten2019SceneTV,Mathew2020DocVQAAD}. However, there is a limited number of hight-qualified and annotated datasets for researching this task in Vietnamese \cite{ViVQA,OpenViVQA,evjvqa,Tran2023ViCLEVRAV,Nguyen2024ViTextVQAAL,Pham2024ViOCRVQANB}.

On the other hand, the main challenge of Text-based VQA is exploiting the meaning of scene texts available in the images so that deep learning methods can recognize them and depend on them to provide the most appropriate answers. They propose to tackle this challenge by introducing several modules \cite{Biten2021LaTrLT,Fang2023SeparateAL,Kil2022PreSTUPF}. However, most of these modules explore the spatial information of scene texts in images via their bounding boxes and their meaning was obtained by using embedding layers from pre-trained language models \cite{Hu2019IterativeAP,Kant2020SpatiallyAM,Gao2020StructuredMA,Biten2021LaTrLT,Fang2023SeparateAL}.

In this study, we inspire the definition of meaning from American Distributionalsm, a field of study in linguistics, and recent works on the Vietnamese lexical system \cite{giap2008,giap2011,caophoneme,chau2007word} to propose a novel method, \textbf{Vi}etnamese \textbf{Cons}tituent Trans\textbf{Former} (ViConsFormer), which effectively incorporate the meaning of Vietnamese scene texts to yield answers. 

Our extensive experiments on the two Text-based VQA datasets in Vietnamese show that our proposed method outperforms previous baselines and proposes several research directions for future studies.

\section{Related works} \label{sec:related-works}

\subsection{Datasets}
Former studies in VQA defined this task as answering questions relevant to objects in images. Antol et al. \cite{VQA} first introduced the VQA task by publishing the VQA dataset.

This novel way of treatment on the VQA dataset results in the language prior phenomenon as indicated by Goyal et al. \cite{Goyal2016MakingTV}. This phenomenon describes that VQA methods tend to yield answers by recognizing the pattern of questions rather than based on evidence from given images.

To overcome the language prior phenomenon, \cite{Goyal2016MakingTV} introduced the VQAv2 dataset. This dataset is the rebalanced version of the VQA dataset constructed by balancing the number of answers belonging to particular patterns of questions.

Making further steps from VQAv2, \cite{singh2019towards} introduced a novel form of VQA task, which is named Text-based VQA tasks in later studies \cite{Hu2019IterativeAP,Biten2021LaTrLT,Li2023BLIP2BL,Fang2023SeparateAL,Kil2022PreSTUPF}. In particular, Text-based VQA tasks require the machines to understand scene texts beside objects in the images and use these scene texts to give the respective answers. Text-based VQA tasks become significantly challenging as the additional modality of scene texts and recently raised attention from many researchers \cite{Hu2019IterativeAP,Biten2021LaTrLT,Li2023BLIP2BL,Fang2023SeparateAL,Kil2022PreSTUPF}.

Although there are numerous VQA datasets in English, there are few VQA datasets in Vietnamese. Tran et al. \cite{ViVQA} first introduced the ViVQA dataset, the first dataset for researching VQA in Vietnamese. Later on, various studies released many datasets, particularly UIT-EVJVQA \cite{evjvqa}, OpenViVQA \cite{OpenViVQA} and ViClever \cite{Tran2023ViCLEVRAV} for VQA task as well as ViTextVQA \cite{Nguyen2024ViTextVQAAL} and ViOCRVQA \cite{Pham2024ViOCRVQANB} for Text-based VQA task.

\subsection{Methods}
Former methods share the same architecture that uses pre-trained CNN-based models to extract image features and RNN-based methods to learn the questions with integrated image features for producing answers \cite{Lu2016HierarchicalQC,Yang2015StackedAN}.

In addition, with the introduction of the attention mechanism \cite{Vaswani2017AttentionIA}, former VQA methods attempted to provide this technique with the assumption of learning the attention relation between images and questions. Typical VQA methods for this approach can be categorized as Co-Attention \cite{Lu2016HierarchicalQC,Yu2019DeepMC}, or Stack Attention \cite{Yang2015StackedAN}.

The development of BERT \cite{Devlin2019BERTPO} provides another architecture that lots of studies inspired as well as introduced numerous novel methods such as \cite{Lu2019ViLBERTPT,Li2019VisualBERTAS,Tan2019LXMERTLC,Su2019VLBERTPO,Zhou2019UnifiedVP,Li2019UnicoderVLAU,Cho2020XLXMERTPC}.

Another way of learning the correlation between images and questions is to define multilinear functions that accept features of questions and images as inputs. Various studies followed this approach and introduced deep learning methods using multilinear functions instead of Transformer \cite{Kim2018BilinearAN,Do2019CompactTI}.

However, most of the VQA methods in English were defined as answer-selection models. Recently, text-based VQA datasets were introduced, and these answer-selection methods can not model effectively Text-based VQA datasets because of the diverse forms of answers. In particular, \cite{OpenViVQA} defined the open-ended VQA task with the publication of the OpenViVQA dataset in Vietnamese. They showed that former VQA methods are challenging to model in this novel form of VQA task. VQA methods using language models are then developed to tackle the challenging of Text-based VQA tasks and open-ended VQA tasks \cite{OpenViVQA}.

The main challenge of the Text-based VQA task is how to model scene texts in images to yield a good answer. Many T5-based VQA methods were introduced with particular modules that try to learn the meaning and spatial relations among scene texts \cite{Biten2021LaTrLT,Kil2022PreSTUPF,Fang2023SeparateAL}

\section{ViConsFormer - Vietnamese Constituent Transformer} \label{sec:proposed-methods}

\subsection{Preliminaries} \label{sec:preliminary}

\subsubsection{Vietnamese Scene Texts}

\begin{figure*}
    \centering
    \includegraphics[width=0.8\textwidth]{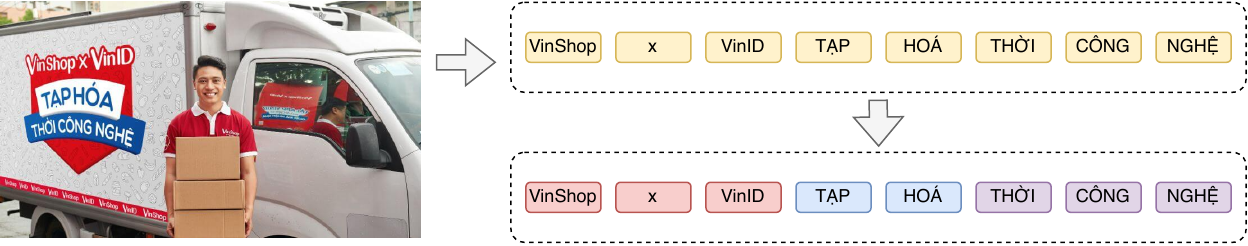}
    \caption{Forming meaningful constituents in the sequence of OCR tokens.}
    \label{fig:meaningful-constituent}
\end{figure*}

Scene texts in images taken in Vietnam have the following rule in general: scene texts on the same line are in the same meaningful constituents. For instance, in Figure \ref{fig:meaningful-constituent}, there are three lines of scene texts on the truck: \textit{VinShop x VinID} indicates the cooperation of the two companies, \textit{Tạp hóa} (grocery) indicates the kind of business of the two companies, and \textit{Thời công nghệ} (technological times) points out the characteristic of the grocery. There does not exist the situation where all scene texts are in the same line, but they are meaningless.

However, current Text-based VQA methods receive scene texts as the line of tokens ordered by the spatial information (the 2D coordinates of bounding boxes) \cite{Biten2021LaTrLT,Fang2023SeparateAL,Kant2020SpatiallyAM,Gao2020StructuredMA,Hu2019IterativeAP}. There is no signal to determine which constituents each scene text token belongs to. SaL \cite{Fang2023SeparateAL} tackled this challenge by providing an additional special token \textit{<context>} between scene text tokens. These tokens learn how to represent the start and end positions of every meaningful constituent.

We approach this challenge in different ways. From our observations, we find that meaningful constituents include complete lexical units, which we call Vietnamese words. To this end, we introduce a novel method that describes the Vietnamese words and hence describes the meaningful constituents of scene text of images taken in Vietnam (Figure \ref{fig:meaningful-constituent}). We continue the in-depth discussion of how to describe words from the line of scene text tokens in the following Section.

\subsubsection{Meaning representation}

Recent studies \cite{Yang2015StackedAN,Biten2021LaTrLT,Gao2020StructuredMA,Kant2020SpatiallyAM} addressed the Text-based VQA task by proposing a module that incorporates the position of scene texts in images via the coordinates of their bounding boxes. However, the meaning of scene texts is not reflected by their spatial positions. One attempt to explore the meaning of scene texts is to use the embedding layer of pre-trained language models such as BERT \cite{Devlin2019BERTPO} and T5 \cite{Raffel2019ExploringTL} as in \cite{Biten2021LaTrLT,Hu2019IterativeAP,Fang2023SeparateAL,Gao2020StructuredMA}. This approach has a limitation in that not all scene texts in images are available in the fixed vocab of the pre-trained language models. When tokens are not in the vocab, pre-trained language models usually segment them into subwords \cite{Wang2019NeuralMT}. This way of representation tends to introduce the ambiguity of scene texts to Text-based VQA methods.

Another approach is constructing a pre-trained model, particularly for scene text representation as in \cite{Kil2022PreSTUPF}. However, training a pre-trained model requires high-cost computational facilities.

In our study, we approach the meaning representation of scene texts following the study of American Distributionalism \cite{bloomfield,harris1951methods} in linguistics. This field of study in linguistics researches language by observing how it appears. They think research in linguistics must be done via observable and measurable units. Linguisticians should avoid falling into unobservable things such as semantics or meanings. They describe languages as the distributions of their constituents, and meaning is defined as the consequence of how words are distributed and how they appear together in sentences. For instance, given the sentence \textit{Everyone in the room knows at least two languages} and the sentence \textit{At least two languages are known by everyone in the room}, these two sentences are different in terms of meaning although they are formed from the same set of words \cite{chomsky2014aspects}.

In addition, Vietnamese lexical structure differs from English. Words in English include one or more than two syllables. While in Vietnamese, words contain one or more syllables, and spaces separate these syllables. For instance, \textit{đại lý} in Vietnamese is a word containing two tokens and two syllables, while in English, it means \textit{agency}, has three syllables and one token. In other words, an English token can be translated into more than one token in Vietnamese.

\begin{figure}
    \centering
    \begin{subfigure}[b]{0.45\textwidth}
        \includegraphics[width=\textwidth]{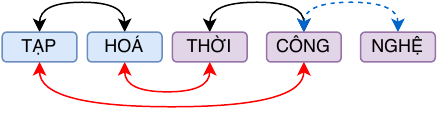}
        \caption{Demonstration of the case where self-attention scores unrelated tokens belonging to different words (the red arrows) but can lack relation among tokens in a word (the dashed arrow).}
        \label{fig:attention-a}
    \end{subfigure}
    \begin{subfigure}[b]{0.45\textwidth}
        \includegraphics[width=\textwidth]{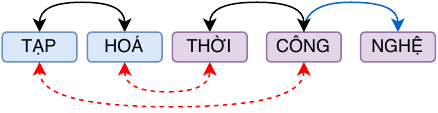}
        \caption{Our Constituent Module was proposed to re-correct the attention scores via the constituent scores (the blue line) while removing the unnecessary relations (dashed red arrows).}
        \label{fig:attention-b}
    \end{subfigure}
    \caption{An example of a noun phrase in Vietnamese. The translated phrase in English: grocery in technological times.}
\end{figure}

\begin{figure*}[htp]
    \centering
    \includegraphics[width=0.8\textwidth]{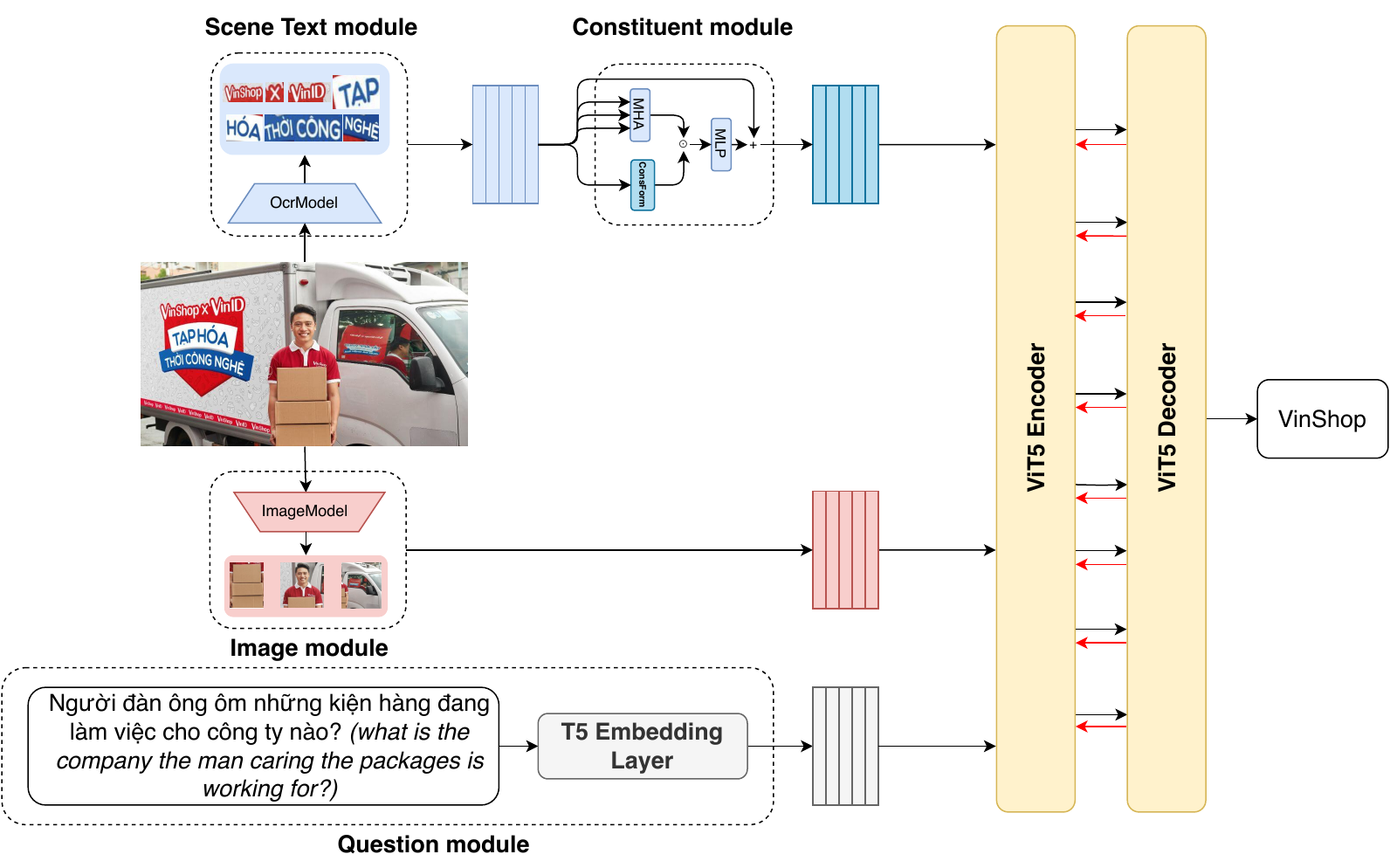}
    \caption{The overall architecture of the ViConsFormer.}
    \label{fig:overall-arch}
\end{figure*}

We follow the recent advantages of the attention mechanism in English. From various studies \cite{Vaswani2017AttentionIA,Bahdanau2014NeuralMT,Luong2015EffectiveAT}, the attention mechanism can describe how words are relevant to each other. However, as analyzed previously, Vietnamese words may include more than one syllable, encoded as tokens in sentences. The attention mechanism has no constraint in how it can form attentive connections. Therefore, in Vietnamese sentences, one token in this word will attend to one token in another, which does not yield any meaning (Figure \ref{fig:attention-a}).

To this end, we introduce the Constituent Module. This module constructs two components: the attention score matrix $\mathcal{A}$ and the constituent score matrix $\mathcal{C}$. The constituent score matrix $\mathcal{C}$ describes the words of scene texts in images by highlighting which tokens belong to a word. 

Moreover, as there are no technical constraints in the attention mechanism, we might have two tokens in different words, but they can be scored to attend to each other. The constituent score matrix $\mathcal{C}$ plays the role of re-correcting the attention score matrix $\mathcal{A}$ so that there are no unnecessary connections among tokens belonging to different words (Figure \ref{fig:attention-b}). The description of how we construct $\mathcal{C}$ was detailed from equation \ref{eq:1} to equation \ref{eq:5}.

\subsection{Overall architecture}

The main contribution of ViConsFormer is the Constituent module. This module is proposed to describe the meaning of scene texts, as discussed in the previous section. In general, our method has five components: Image Embedding module, Question Embedding module, Scene Text Embedding module, Constituent module, and Multimodal backbone (Figure \ref{fig:overall-arch}).

\subsubsection{Constituent module}

The Constituent module includes two components: the multi-head attention \cite{Vaswani2017AttentionIA} determining attention score $\mathcal{A}$ and Constituent formation determining constituent score $\mathcal{C}$.

In Vietnamese morphology, syllables in words have two kinds of semantic relations \cite{giap2008,giap2011}:
\begin{itemize}
    \item Syllables in a word contribute their meanings equally to the overall meaning of that word. For instance, \textit{quần áo} means clothes in general, compounding \textit{quần} (paints in general) and \textit{áo} (shirts in general).
    \item One syllable defines the core meaning of the word, and other syllables play the role of modifiers. These modifiers narrow down the meaning of the main syllable so that the meaning of the whole word is more particular. For instance, \textit{nhà ăn} (cafeteria) includes syllable \textit{nhà} (houses in general) and \textit{ăn} (dining in general).
\end{itemize}

Given the sequence of scene texts $f_{ocr} = (f^{ocr}_1, f^{ocr}_2, ..., f^{ocr}_n)$ obtained from the embedding layer of ViT5 \cite{Phan2022ViT5PT} as input, we model these kinds of semantic relations by defining a bilinear function:
\begin{equation} \label{eq:1}
    r_{k, k+1} = f(f^{ocr}_k, f^{ocr}_{k+1}) = f^{ocr}_k W (f^{ocr}_{k+1})^T
\end{equation}
where $W$ is the learnable parameter. Then with every token $ith$, we describe the semantic relations with its neighbor tokens $(i-1)th$ and $(i+1)th$ as:
\begin{equation} \label{eq:2}
    pr_{i-1, i}, pr_{i, i+1} = softmax(r_{i-1, i}, r_{i, i+1})
\end{equation}

If token $ith$ has semantic relations with token $(i-1)th$, and token $(i-1)th$ is in another word, then we expect $pr_{i-1, i} > pr_{i, i+1}$ (and vice versa). In the case token $ith$ has semantic relations with both token $(i-1)th$ and $(i+1)th$, the mass of $pr_{i-1, i}$ and $pr_{i, i+1}$ determine how much relevant these tokens share (contributing equally to the overall meaning, or main-secondary meaning contribution, or there is no connection among these tokens).

In addition, as the consequence of the asymmetry of matrix multiplication, we have $pr_{k, k+1} \ne pr_{k+1, k}$ while they describe the same idea. To this end, we define the probability $P_k$ to measure the semantic relations of token $k$ with its neighbor tokens. $P_k$ is obtained by averaging over $pr_{k, k+1}$ and $pr_{k+1, k}$:
\begin{equation} \label{eq:3}
    P_k = \sqrt{pr_{k, k+1} \times pr_{k+1, k}}
\end{equation}

Defining $\mathcal{C}_{ij}$ the probability of "tokens from the position $ith$ to the position $jth$ are in the same constituent", together with the definition of $P_k$, we have:
\begin{equation} \label{eq:4}
    \mathcal{C}_{ij} = \prod_{k=i}^{j-1}{P_{k}}
\end{equation}

It is worth noting that $P_k \in [0, 1]$, hence $\mathcal{C}_{ij}$ rapidly converges to $0$ when $k \rightarrow \infty$, which results in the gradient vanishing. To avoid this problem, we re-formulate $\mathcal{C}_{ij}$ as:
\begin{equation} \label{eq:5}
    \begin{split}
        \mathcal{C}_{ij} & = exp(log(\mathcal{C}_{ij})) \\
        & = exp\left(log\left(\prod_{k=i}^{j-1}{P_k}\right)\right) \\
        & = exp\left(\sum_{k=i}^{j-1}{log(P_k})\right)
    \end{split}
\end{equation}

To construct the attention score matrix $\mathcal{A}$, we perform the self-attention by defining the query $Q$, key $K$, and value $V$ as:
$$
    Q  = W_q f_{ocr}
$$

$$
    K  = W_k f_{ocr}
$$

$$
    V  = W_v f_{ocr}
$$
where $W_q, W_k, W_v \in \mathbb{R}^{d_{model} \times d_{model}}$ are learnable parameters.

The attention scores over detected scene texts are determined as follows:

$$
    \mathcal{A} = softmax\big(\frac{QK^T}{\sqrt{d_{model}}}\big)
$$

The final attention score matrix is determined by providing $\mathcal{A}$ with the constituent score matrix $\mathcal{C}$:
\begin{equation} \label{eq:6}
    \mathcal{S} = \mathcal{A} \odot \mathcal{C}
\end{equation}

As both $\mathcal{C}$ and $\mathcal{A}$ have the form of exponential functions, it is worth noting that the constituent scores are not added to the attention scores intuitively by summation but by element-wise multiplication.

Finally, the output of scene text features is determined as follows:
\begin{equation} \label{eq:7}
    f^{ocr}_{out} = \mathcal{S} V
\end{equation}

\subsection{Image Embedding module}

We follow the module for object features representation of SaL to represent the features of images. In particular, the features of objects $f_{obj} = (f^{obj}_1, f^{obj}_2, ..., f^{obj}_n)$ available in images are determined as follows:
\begin{equation}
    \begin{split}
        f_i^{obj} & = ViT5LN(W_{fr}^{obj} x_i^{obj, fr}) \\
        & + ViT5LN(W_{bx}^{obj} x_i^{obj, bx})
    \end{split}
\end{equation}
where $W_{fr}^{obj}$ and $ W_{bx}^{obj}$ are trainable parameters, $ViT5LN$ is the normalization layer of ViT5, and $x^{obj, fr}_i$ is the features of object $ith$ in image.

Unlike SaL, we do not consider the features of object tags, as their appearance in our proposed methods does not yield any significant improvement in scores. We will provide more numerical information about this statement in Section \ref{sec:ablation}.

\subsection{Scene Text Embedding module}

To obtain features of scene texts in images, we follow SaL to determine these features:
\begin{equation}
    \begin{split}
        f_i^{ocr} & = ViT5LN(W_{fr}^{ocr} x_i^{ocr, fr})\\
        & + ViT5LN(W_{bx}^{ocr} x_i^{ocr, bx}) \\
        & + W_{ViT5}^{ocr} x_i^{ViT5}
    \end{split}
\end{equation}
where $W_{fr}^{ocr}$, $ W_{bx}^{ocr}$, and $W_{ViT5}^{ocr}$ are trainable parameters, $ViT5LN$ is the normalization layer of ViT5, $x^{obj, fr}_i$ is the features of object $ith$ in image, and $x_i^{ViT5}$ is the text embedding of scene text token $ith$ produced by the embedding layer of ViT5.

\subsection{Question Embedding module}
Following LaTr \cite{Biten2021LaTrLT}, the questions are embedded into $f_Q = (q_1, q_2, ..., q_L)$ using the embedding layer of ViT5 where each position $q_i \in \mathbb{R}^d$.

\subsection{Multimodal backbone}

Following previous works \cite{Biten2021LaTrLT,Fang2023SeparateAL,Kil2022PreSTUPF}, we use the T5-based pre-trained model as the multimodal backbone. However, as our experiments were conducted on the Vietnamese Text-based VQA dataset, we provide ViConsFormer with ViT5 \cite{Phan2022ViT5PT} pre-trained language models.

The input to the ViT5 backbone is defined as the fused features $f_f$ constructed by concatenating $f_{obj}$, $f_{ocr}$, and $f_Q$:

$$
    f_f = [f_{obj};f_{ocr}; f_Q]
$$

\section{Experiments} \label{sec:experiments}

\subsection{Datasets}
In this work, we evaluate our proposed methods on the two datasets ViTextVQA \cite{Nguyen2024ViTextVQAAL} and ViOCRVQA \cite{Pham2024ViOCRVQANB}.

The ViTextVQA dataset was constructed by asking questions and answering answers relevant to scenario images. These images were street views in Viet Nam \cite{Nguyen2024ViTextVQAAL}. Scene texts in this dataset are diverse in positions, colors, light conditions, transformations, shapes, and meaning. As indicated in \cite{Nguyen2024ViTextVQAAL}, the smaller size of scene texts in the images lead to more challenges in producing answers.

The ViOCRVQA dataset was constructed semi-automatically by collecting book covers from websites \cite{Pham2024ViOCRVQANB}. The authors built question templates, then filled in these templates and extracted answers via the corresponding metadata of the books.

\subsection{Metrics}
We follow the experiments on the two datasets ViTextVQA \cite{Nguyen2024ViTextVQAAL} and ViOCRVQA \cite{Pham2024ViOCRVQANB} to use the Exact Match (EM) and F1-token as main metrics in our evaluation.

Accordingly, let $P = \{p_1, ..., p_m\}$ is the predicted answers, and $G = \{g_1, ..., g_n\}$ is the truth answers. The M of each predicted-truth answer is determined as follows:
$$
    EM = \delta_{P, G}
$$
where $\delta_{x, y}$ is the Kronecker symbols which $\delta_{x,y} = 1$ when $x = y$ and 0 otherwise.

The F1-Token metric is defined as the harmonic mean of the Precision and Recall (in token level) as:
$$
    Pr = \frac{P \cap G}{P}
$$

$$
    Re = \frac{P \cap G}{G}
$$

$$
    \text{F1-Token} = \frac{2 Pr Re}{Pr + Re}
$$

The overall EM and F1-Token are averaged over all predicted-truth answers of the whole dataset.

\subsection{Configuration}

In our experiment, we trained the ViConsFormer following the previous studies on ViTextVQA and ViOCRVQA datasets \cite{Nguyen2024ViTextVQAAL,Pham2024ViOCRVQANB} that used ViT5 \cite{Phan2022ViT5PT} as the multimodal backbone. For the $ImageModel$ we deployed the VinVL \cite{Zhang2021VinVLMV} pre-trained image models. We used SwinTextSpotter \cite{Huang2022SwinTextSpotterST} to obtain Vietnamese scene texts from images to extract their detection features and recognition features. The ViConsFormer was trained in a single run, using Adam \cite{Kingma2014AdamAM} as optimizer on an NVIDIA A100 80GB GPU. The batch size was set to 32 and the learning was set to $1e^{-4}$. We applied the early stopping technique to train ViConsFormer.

\subsection{Baselines}

To evaluate the effectiveness of our proposed ViConsFormer on the Vietnamese Text-based VQA dataset, we compared this method with the following baselines:
\begin{itemize}
    \item M4C \cite{Hu2019IterativeAP}: M4C is the first vision-language learning task that was constructed based on the Transformer architecture \cite{Vaswani2017AttentionIA}. Its multimodal backbone is BERT \cite{Devlin2019BERTPO}. M4C approaches the text-based VQA task by sequentially generating tokens to form the answers. Tokens of answers can be obtained from the vocabulary or copied from the scene texts available in the images using the Pointer Network \cite{Hu2019IterativeAP} module.
    \item LaTr \cite{Biten2021LaTrLT}: This is the first method that integrated spatial information of scene texts into the multimodal backbone. They encoded the coordinates of the bounding boxes into 4-dimensional vector space, then projected them directly to the latent space of the multimodal backbone and added them to the features of scene texts. Unlike M4C, LaTr proposed using T5 \cite{Raffel2019ExploringTL} as its multimodal backbone and using a subword tokenizer to encode scene texts. Scene texts in the images that are not available in the vocabulary of the T5 pre-trained model are subsegmented into sequences of chunks. Hence, instead of copying scene texts from images via a particular module, it learns how to form out-of-vocabulary scene texts from respective subwords.
    \item PreSTU \cite{Kil2022PreSTUPF}: Instead of modeling the relation among scene texts via their spatial relations, PreSTU pre-trained the T5 backbone to approximate the distribution of the scene texts. The particular technique of PreSTU differs from other methods in that they sort the scene texts in left-right top-bottom orders.
    \item SaL \cite{Fang2023SeparateAL}: SaL proposed integrating the labels of objects and tokens of scene texts to their respective visual features. These labels and tokens are embedded by the embedding layer of the T5 backbone to yield the textual meaning of the objects and scene texts. Moreover, instead of encoding the coordinates of bounding boxes, they introduce another way, which is to measure the relative position among scene texts in the images.
    \item BLIP-2 \cite{Li2023BLIP2BL}: BLIP-2 proposed the Q-Former module, which is fine-tuned to connect the latent space between two frozen pre-trained models: pre-trained language model and pre-trained image model. This method was pre-trained using three objective functions: Image-Text matching, Image-Text Contrastive learning, and Image-grounded Text generation. The adaption of BLIP-2 is to fine-tune the Q-Former on the downstream tasks while keeping the pre-trained image and language models frozen.
\end{itemize}

\subsection{Results}

\begin{table}[htp]
    \centering
    \caption{Main results of the ViConsFormer and the baselines on the ViTextVQA and ViOCRVQA datasets. The scores of baselines are obtained from previous studies \cite{Nguyen2024ViTextVQAAL,Pham2024ViOCRVQANB}.}
    \label{tab:main-results}
    \resizebox{0.45\textwidth}{!}{
    \begin{tabular}{cccccc}
        \hline
         &  & \multicolumn{2}{c}{\textbf{ViTextVQA}} & \multicolumn{2}{c}{\textbf{ViOCRVQA}} \\ \cline{3-6} 
        \multirow{-2}{*}{\textbf{\#}} & \multirow{-2}{*}{\textbf{Method}} & \textbf{F1-token} & \textbf{EM} & \textbf{F1-token} & \textbf{EM} \\ \hline
        1 & M4C & 30.04 & 11.60 & - & - \\
        2 & BLIP2 & 37.78 & 15.01 & 55.23 & 21.45 \\
        3 & LaTr & 43.13 & 20.42 & 60.97 & 30.80 \\
        4 & PreSTU & 43.81 & 20.85 & 66.25 & 33.86 \\
        5 & SAL & 44.89 & 20.97 & 67.25 & \textbf{39.08} \\ \hline
        6 & ViConsFormer (ours) & {\color[HTML]{333333} \textbf{45.58}} & {\color[HTML]{333333} \textbf{22.72}} & {\color[HTML]{333333} \textbf{70.92}} & {\color[HTML]{333333} 37.65} \\ \hline
    \end{tabular}}
\end{table}

In general, the evaluation scores on the ViTextVQA dataset are lower than those on the ViOCRVQA dataset. This can be explained by the questions in the ViTextVQA dataset being annotated manually by Vietnamese native speakers, while questions in the ViOCRVQA were constructed semi-automatically using constructed templates. Therefore, the patterns of questions in the ViOCRVQA dataset are easier to explore. In addition, images from the ViTextVQA dataset are scenery views in Viet Nam, including street signs, signboards, addresses, banners, places, etc. Scene texts available in images from ViTextVQA are complicated under various transformations, colors, light conditions, and sizes and are relevant to diverse objects. In the ViOCRVQA dataset, scene texts are more clarified and belong to particular categories such as titles, names of authors, publishers, and translators \cite{Pham2024ViOCRVQANB}.

On the ViTextVQA dataset, our proposed methods decisively outperformed all the given baselines. In particular, M4C using BERT as its multimodal backbone yielded the lowest scores. Text-based VQA methods using T5 as their multimodal VQA backbone significantly achieved higher scores.

On the ViOCRVQA, our method significantly outperforms all the baselines on the F1-Token metric. However, on EM, our method drops down its score compared to SaL.

\subsection{Ablation study} \label{sec:ablation}

\subsubsection{Ablation study on $\mathcal{A}$ and $\mathcal{C}$}

\begin{table}[htp]
    \centering
    \caption{Ablation study on attention score matrix $\mathcal{A}$ and constituent score matrix $\mathcal{C}$}
    \label{tab:ablation-matrices}
    \begin{tabular}{l|c|c|c|c}
        \hline
        \textbf{Dataset} & $\mathcal{A}$ & $\mathcal{C}$ & \textbf{F1-token} & \textbf{EM} \\ \hline
        \multirow{3}{*}{ViTextVQA} & \cmark & \xmark & 0.4309 & 0.2038 \\
         & \xmark & \cmark & 0.4025 & 0.1740 \\
         & \cmark & \cmark & \textbf{0.4558} & \textbf{0.2272} \\ \hline
        \multirow{3}{*}{ViOCRVQA} & \cmark & \xmark & 0.6503 & 0.3204 \\
         & \xmark & \cmark & 0.6172 & 0.3132 \\
         & \cmark & \cmark & \textbf{0.7092} & \textbf{0.3765} \\ \hline
    \end{tabular}
\end{table}

The Constituent module determines two matrices: the attention score matrix $\mathcal{A}$ and the constituent score matrix $\mathcal{C}$. We expect the constituent score matrix will re-correct the unnecessary relations scored by the attention score matrix to represent the meaning of scene texts in images appropriately. To show how these two matrices interact with each other, we conduct experiments in case only the attention score matrix is calculated, and only the constituent score matrix is calculated.

According to Table \ref{tab:ablation-matrices}, the Constituent module with only attention score matrix $\mathcal{A}$ performed better than when being replaced by the constituent score matrix $\mathcal{C}$. However, having the constituent score matrix $\mathcal{C}$ to re-correct the attention score matrix $\mathcal{A}$ leads to significant improvement in both metrics.

\subsubsection{The necessity of object labels}

\begin{table}[htp]
    \centering
    \caption{Ablation study of the ViConsFormer on the Scene Text module and Image  module. \textit{Labels} indicate the labels of detected objects and \textit{Tokens} indicate the tokens of detected scene texts.}
    \label{tab:ablation}
    \resizebox{0.45\textwidth}{!}{
    \begin{tabular}{c|c|c|c|c|c}
        \hline
         &  & \multicolumn{2}{c|}{\textbf{Labels}} & \multicolumn{2}{c}{\textbf{Tokens}} \\ \cline{3-6} 
        \multirow{-2}{*}{\textbf{Dataset}} & \multirow{-2}{*}{\textbf{Metrics}} & \textbf{\ding{52}} & \textbf{\ding{56}} & \textbf{\ding{52}} & \textbf{\ding{56}} \\ \hline
         & F1-Token & { 45.58 } & {{\color{red} $\downarrow$} 0.25} & { 45.58 } & {{\color{red} $\downarrow$} 1.22} \\
        \multirow{-2}{*}{ViTextVQA} & EM & { 22.72 } & {{\color{red} $\downarrow$} 0.40} & { 22.72 } & {{\color{red} $\downarrow$} 1.19} \\ \hline
         & F1-Token & { 70.92 } & {{\color{blue} $\uparrow$} 0.12} & { 70.92 } & {{\color{red} $\downarrow$} 2.55} \\
        \multirow{-2}{*}{ViOCRVQA} & EM & { 37.65 } & {{\color{red} $\downarrow$} 0.52} & { 37.65 } & {{\color{red} $\downarrow$} 1.60} \\ \hline
    \end{tabular}}
\end{table}

As mentioned in Section \ref{sec:proposed-methods}, in the Image module of ViConsFormer, the labels of detected objects are not required but the tokens of scene texts. To show this claim, we conducted an ablation study for ViConsFormer on the ViTextVQA and ViOCRVQA datasets.

As indicated in Table \ref{tab:ablation}, there is no significant performance degradation in both F1-Token and EM scores if we do not provide ViConsFormer with object labels. However, ViConsFormer obtained significantly lower scores when it did not see scene text tokens. These results indicate that not the labels of detected objects but tokens of detected scene texts influence the overall performance.

\section{Conclusion}

In this study, we introduced the Constituent module. Hence, the ViConsFormer, inspired by the SaL method, approaches the main challenge of Text-based VQA in Vietnamese in a novel way. Experimental results indicate that our proposed method is effective on both Text-based VQA datasets.

\section{Limitations}

Although our ViConsFormer addresses the challenge of Text-based VQA task by proposing the Constituent module, this method has some limitations that need to be improved and studied in the following studies.

The first limitation is our assumption of linearity while modeling the semantic relationship between two continuous scene text tokens. This assumption is proposed for the simplicity in our novel method. It is necessary to explore the form of this semantic relationship and find the appropriate ways of modeling it in subsequent studies.

The second limitation is that we give a naive treatment for the fused futures $f_f$ when passing them forward to the multimodal backbone. There are various ways of obtaining these fused features, such as using the Co-Attention mechanism \cite{Yu2019DeepMC,Lu2016HierarchicalQC,Yang2015StackedAN}, or multilinear functions \cite{Do2019CompactTI,Kim2018BilinearAN}. We will leave these directions in our future work.

\section{Acknowledgement}
This research is funded by Vietnam National University HoChiMinh City (VNU-HCM) under the grant number DS2024-26-01.

\bibliography{custom}




\end{document}